\documentclass[sigconf]{acmart}
\usepackage{xcolor}
\usepackage{caption}
\usepackage{siunitx}

\AtBeginDocument{%
  }

\setcopyright{acmlicensed}
\copyrightyear{2024}
\acmYear{2024}

\acmConference[HRI '24]{ ACM/IEEE International
Conference on Human Robot Interaction (HRI)}{March 11-15, 2024}{Boulder, CO, USA}

\begin{document}

\title{Prosody for Intuitive Robotic Interface Design: \\It's Not What You Said, It's \underline{How} You Said It}

\author{Elaheh Sanoubari$^1$, Atil Iscen$^2$, Leila Takayama$^3$, Stefano Saliceti$^2$, \\Corbin Cunningham$^1$ and Ken Caluwaerts$^2$\\
\small{\textit{$^1$Google Labs $^2$Google DeepMind $^3$Hoku Labs}.
\texttt{\{eqs, atil, ssaliceti, corbinc, kencaluwaerts\}@google.com, takayama@hokulabs.com}}}

\renewcommand{\shortauthors}{Sanoubari et al.}

\begin{abstract}
In this paper, we investigate the use of `prosody' (the musical elements of speech) as a communicative signal for intuitive human-robot interaction interfaces. Our approach, rooted in Research through Design (RtD), examines the application of prosody in directing a quadruped robot navigation. We involved ten team members in an experiment to command a robot through an obstacle course using natural interaction. A human operator, serving as the robot's sensory and processing proxy, translated human communication into a basic set of navigation commands, effectively simulating an intuitive interface. During our analysis of interaction videos, when lexical and visual cues proved insufficient for accurate command interpretation, we turned to non-verbal auditory cues. Qualitative evidence suggests that participants intuitively relied on prosody to control robot navigation. We highlight specific distinct prosodic constructs that emerged from this preliminary exploration and discuss their pragmatic functions. This work contributes a discussion on the broader potential of prosody as a multifunctional communicative signal for designing future intuitive robotic interfaces, enabling lifelong learning and personalization in human-robot interaction.

\end{abstract}


\maketitle

\section{Introduction}
In 1968, Douglas Engelbart's 'mother of all demos' showcased the first Graphical User Interface (GUI), revolutionizing human-machine interaction and shaping the future of computers for decades to come. Engelbart posed this question, \textit{``If in your office, you as an intellectual worker were supplied with a computer display, backed up by a computer that was alive for you all day, and was instantly responsive to every action you had, how much value would you derive from that?''} We extend the inquiry: What if the responsive computer was a robot? An important distinction that sets Engelbart's computer apart from a robot is that the latter is a mobile agent capable of navigating through our physical space. Understanding and following navigation commands is paramount to a mobile robot that is instantly responsive to humans. Motivated by this, we set out to explore how we might design an intuitive control interface that allows a co-located human to guide a robot's navigation. Adopting a Research through Design (RtD) approach, we invited participants to guide a quadruped robot through an obstacle course using any communication that \textit{felt natural} to them. To simulate an intuitive interface, a human 'wizard' acted as the robot's sensory and cognitive proxy, interpreting and executing participants' commands. 

In our analysis of interaction videos, we observed that prosodic cues (musical elements of speech) played a crucial role in accurately interpreting commands when lexical and visual cues were inadequate. These cues, found unexpectedly, supplied essential context to disambiguate numerous commands. Although the wizard could discern and react to these cues while operating the robot, they were not apparent in the transcribed commands. We discuss related work and present an overview of the concept of prosody in section \ref{rw}, describe our methodology in section \ref{methodology}, and present an overview of specific prosodic patterns that we found evidence for in section \ref{findings}.

Prosody is a rich communicative signal that carries information relevant to the task of controlling a co-located robot's navigation, and as such, it holds promise for influencing the design of intuitive robotic interfaces. While Natural Language Understanding (NLU) for robot control is gaining interest, prosody can complement this by providing supplementary context that is difficult or inefficient to express through spoken language alone. Prosody's multifunctional nature and the nuanced control people have over it make it an appealing signal for effective and user-friendly robotic control interfaces. Computational systems that objectively capture prosodic cues via acoustic analysis of speech signal can outperform humans \cite{ward19_interspeech}. We extend this discussion in section \ref{discussion}.

Finally, prosody's \textit{pragmatic} and \textit{paralinguistic} functions make it relevant to lifelong learning and personalization in human-robot interaction (HRI). Pragmatic functions of prosody, such as directing attention, expressing uncertainty, prioritizing information, and coordinating actions are crucial aspects for effective HRI and life-long in-context learning. Additionally, prosody conveys paralinguistic information, such as user traits (e.g., age), emotions (e.g., anger), and cognitive states (e.g., tiredness), which are valuable in many robotic applications, and can facilitate speaker identification and personalized HRI experiences. As a preliminary exploration of prosody in HRI, this work aims to promote utilizing prosodic cues for designing novel intuitive robotic control interfaces.

\section{Background \& Related Work}
\label{rw}
To facilitate effective HRI, we need interfaces that enable seamless bidirectional human-robot communication. Traditionally, research in human-robot communication has concentrated on strictly-defined problem domains. This has involved examining the effects of specific communication modalities (such as `gaze') using the robot either as a stimulus to assess the limits of human reactions, or as a means to explore design variations by altering one modal aspect at a time (e.g., the `duration' of gaze) and gauging its impact on human behavior \cite{admoni2017social}. Despite robots performing well in demos involving a single communication medium, such performance is limited in that it relies on stringent constraints and tight coupling of both the environment and the user \cite{marge2022spoken}.

Natural human interaction involves a great deal of redundancy, as the same message is typically communicated through multiple channels such as words, prosody, gaze, posture, facial expressions, hand gestures, and actions \cite{admoni2017social, marge2022spoken, gaschler2012modelling, ward2019prosodic, gaschler2012social}. We advocate for moving towards holistic and intuitive robotic interfaces designed to align with natural human interactions. We define an intuitive interface for human-robot interaction to be an integrated multi-modal communication system that satisfies two general requirements:
\begin{enumerate}
    \item Has competencies that allow it to: read embodied signals, generate robot messages, and transmit them to humans effectively such that they are immediately and easily understood.
    \item Has affordances that allow it to: receive system-directed messages via natural human interaction, interpret them to extract action-relevant directives, and facilitate action. 
\end{enumerate}

Along this line, building robot controllers based on spoken interaction is gaining traction \cite{marge2022spoken}. Prior work that has outlined 25 recommendations as key advances needed for effective spoken interaction with robots, emphasizes the requirement to `better exploit prosodic information' \cite{marge2022spoken}. While speech, encompasses both words and prosody, traditional speech recognition systems primarily focus on words, often overlooking the rich information embedded in prosody. While this omission may be inconsequential in certain applications, such as asking a voice assistant to play music, it becomes more significant in the context of robots, given that they are mobile agents embodied in our space. In certain HRI instances, what is conveyed via prosody can be more important than the the actual words spoken. As an example, Marge et al. \cite{marge2022spoken} point to how the word \textit{``oops''} can vary in meaning from a `minor mistake' to a `significant surprise', with prosody being the key indicator of the severity of the situation.

Prosody, the musical aspect of speech, encompasses elements such as fundamental frequency (F0), loudness, timing, and spectrual information. Prosody is pervasive in everyday life and plays a vital role in communication by conveying information that extends beyond the literal meaning of words. \textit{Prosodic constructs}, defined as `temporal configurations of prosodic features that carry specific meanings'  \cite{ward2019prosodic} are complex, but systemic. Prosodic constructs transcend direct alignment with words, can vary in degree, and can be superimposed on other prosodic features to convey different meanings or nuances \cite{ward19_interspeech}. Effective use of prosody can significantly enhance communication and its lack can lead to confusion or delayed responses from interaction partners.

Prosody serves many pragmatic functions; for example, we use specific configurations of prosodic elements for directing attention (\textit{``over there!''}), conveying uncertainty (\textit{``the round one?''}), establishing priorities (\textit{``help!''}), and coordinating action (\textit{``three, two, one… go!''}) \cite{marge2022spoken}. Moreover, English prosody includes specific constructs that distinguish different \textit{dialogical activities}, such as explaining, arguing, or making decisions \cite{ward2019prosodic}. Such prosodic features can be objectively extracted via accoustic processing of speech signals (e.g., \cite{eyben2010towards, schuller2013computational}). In fact computational models can already outperform humans in detecting certain prosodic cues. For example, Skantze has proposed a model using LSTM Recurrent Neural Networks can predict whether a conversational turn-shift will occur or not in pauses, better than human observers, using a fairly basic set of five prosodic features \cite{skantze2017towards}.
The ability to extract prosodic information from human communication signals is not merely beneficial for long-term HRI, but it is rather essential for effectively organizing in-context continual robot learning.

\section{Methodology}
\label{methodology}
To explore intuitive robotic interfaces, we took a Research through Design (RtD) approach: a methodology where design practice is used as a means to conduct research, which may involve iterative designing and prototyping to explore complex problems, generate `tacit knowledge', and propose innovative solutions \cite{lupetti2021designerly, mascitelli2000experience}.

\begin{figure}[htb]
\centering
\includegraphics[width=0.4\textwidth]{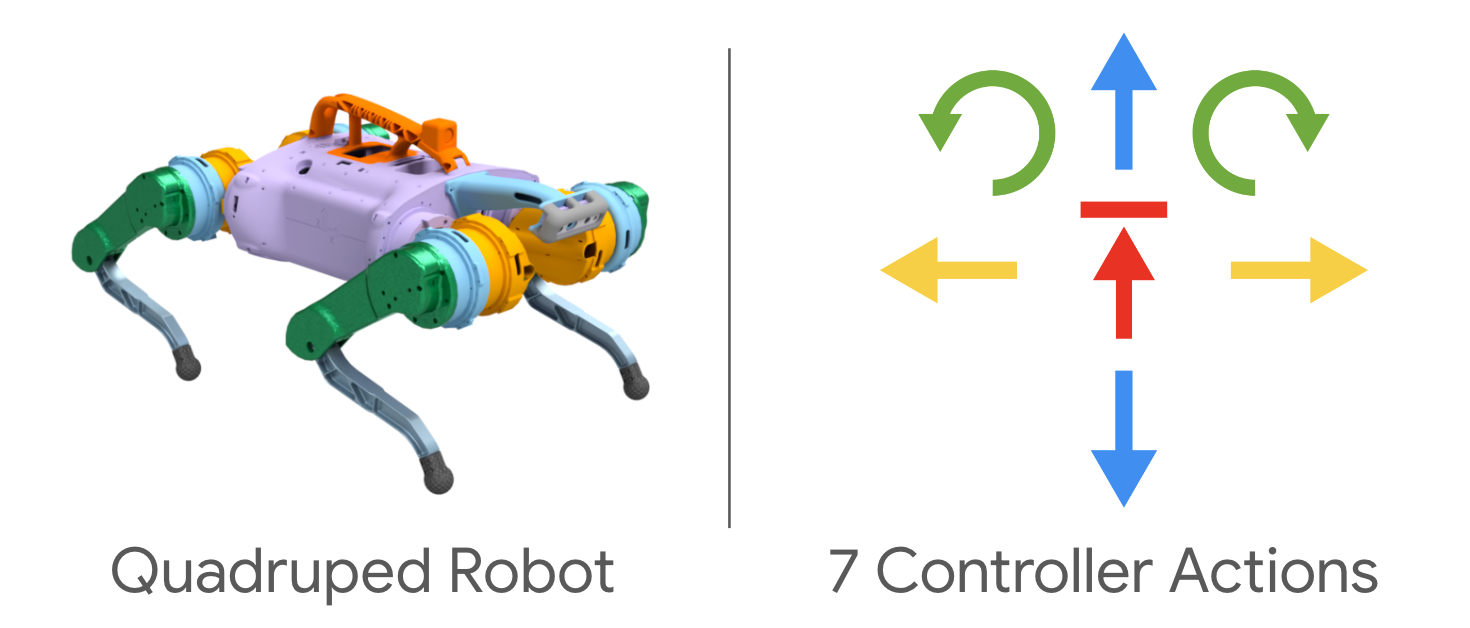}
\caption{Robot and the controller actions.}
\vspace{-1.5em}
\label{controller}
\end{figure}

The robot used in this study was a small quadruped robot developed in-house, that weighs approximately \SI{13.5}{\kilo\gram}, and stands at approximately \SI{0.4}{\meter} tall (see Fig. \ref{controller}). To prototype an intuitive interface, we invited human participants to guide a co-located quadruped robot through an obstacle course, while a human operator \textit{wizarded} \cite{riek2012wizard} the robot by interpreting human instructions in real-time and translating them into basic robot navigation commands. The navigation controller used by the wizard was a command-line interface that facilitated interaction through keyboard inputs, enabling seven distinct actions: moving forward, backward, left, right; turning left or right; and stopping (see Fig. \ref{controller}). 

\subsection{Procedure}
\begin{figure}[htb]
\centering
\includegraphics[width=0.4\textwidth]{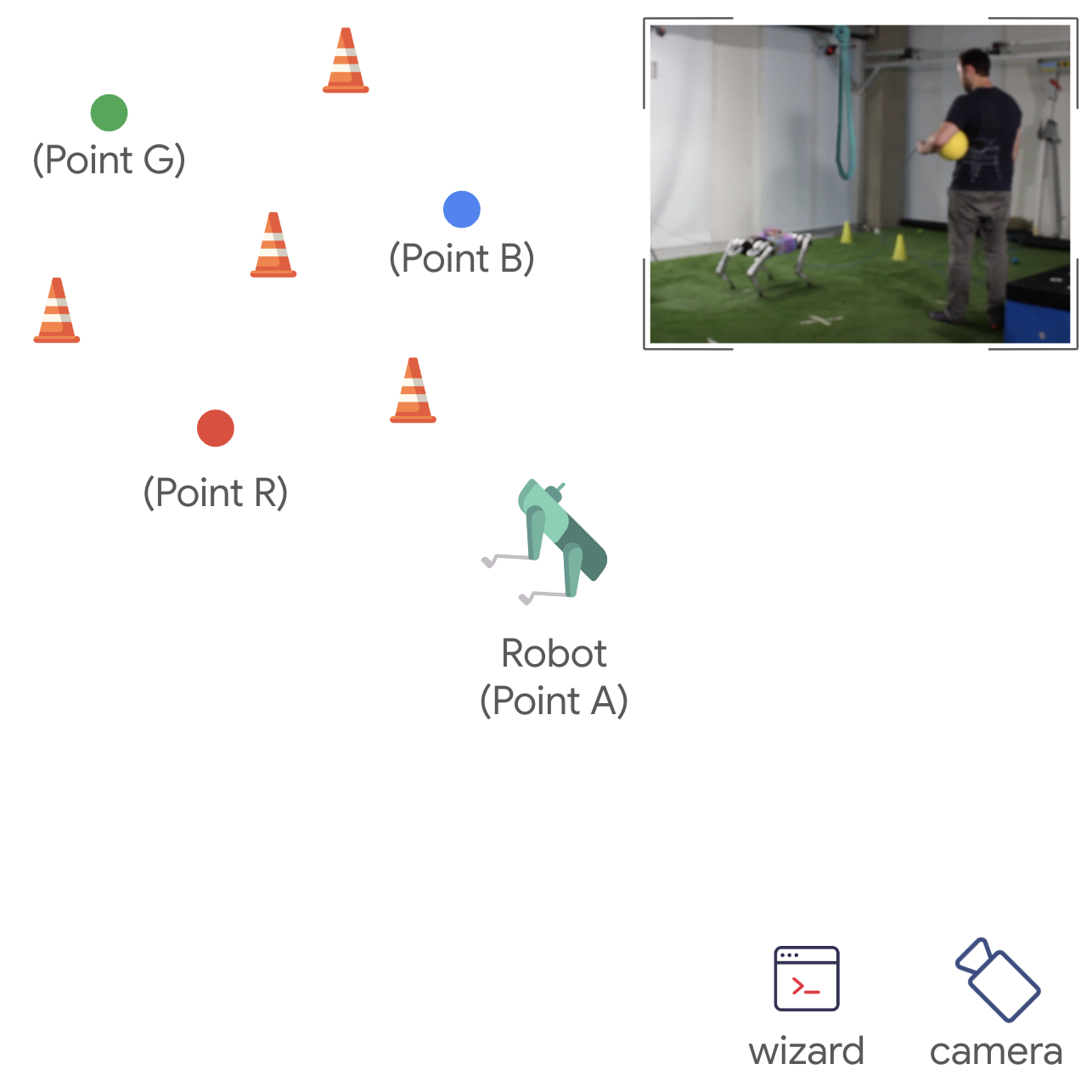}
\caption{Experimental Setup, and a snapshot from the study}
\label{setup}
\vspace{-1.5em}
\end{figure}

We set up the experiment room as depicted in Fig. \ref{setup}. Participants were told the goal of this study is to explore how people might give a quadruped robot navigation commands via natural human interaction. We asked them to guide the robot through the obstacle course by giving it instructions in any way that felt natural to them. Specifically, we asked participants to navigate the robot (initially placed at the center of the room, point A) to the three balls in RGB order: first the red ball (point R), then the green ball (point G), and finally the blue ball (point B); then, navigate back to the center of the room (point A), while avoiding the obstacle cones.

\subsection{Participants and Data}
We invited 10 individuals with moderate to high familiarity with the robot who regularly operated quadrupeds (on a daily to weekly basis). Participants had varying job titles, such as research scientist, roboticist, robotics engineer, robot operator, and software engineer. The collected interaction data amounted to 1.5 hours of recorded video, which was transcribed into a total of 194 verbal commands.

\section{Analysis}
We conducted qualitative analysis by identifying and organizing patterns in the data through thematic analysis \cite{braun2006using} and grouping related ideas using affinity diagramming \cite{holtzblatt2004rapid}. Given our primary goal of understanding the minimum technical requirements for designing an interface that could replace the human `wizard', our initial approach was to analyze the `transcripts' of user commands. However, we quickly realized that the majority of the transcribed verbal commands were inherently ambiguous; that is, it was challenging to map specific lexical commands such as \textit{'go around'}, to precise controller actions (as depicted in Fig. \ref{controller}). 

Our initial attempt to address command ambiguity involved incorporating visual cues. While this improved overall clarity, it did not resolve the issue; for instance, the word `nice' was used in varied contexts, sometimes indicating `keep going', other times indicating `stop', rendering both lexical and visual cues ineffective in disambiguating the intended meaning. 
Further examination of these ambiguous cases revealed a key insight: the distinguishing factor often lay in the presence of different prosodic cues in the audio of the speech. This led to the realization that prosodic cues were essential for accurate command interpretation in many cases.

\section{Observations: Prosody for HRI}
\label{findings}
Below, we introduce a few of the specific prosodic constructs identified in our qualitative analysis, elaborate their pragmatic functions, and provide examples of observed qualitative evidence.

\subsection{High-Priority Interpolation Construction}
This prosodic pattern, characterized by a slow rise in pitch, low intensity, and a fast speaking rate, focuses on marking utterances that deviate from the normal topic structure, such as priority topics (i.e., things that must be done immediately). Additional features often present include a subsequent silence and breathy voice (see chapter 12 \cite{ward2019prosodic}). In this study, we observed that participants frequently employed this prosodic construct to `convey urgency' in their commands. Exemplifying the use of this construct, in one interaction instance, as the robot approached a cone, initially, the participant calmly instructed the robot, \textit{`turn a bit to the left and stop...'}; however, as the robot took another step and dangerously neared the cone, the participant's repeated and hastened \textit{`\underline{left-left-left-left!}'} This served as a prosodic cue to convey a `shift in urgency' and signal to the robot the critical need to turn left immediately. 

\subsection{Minor Third Construction}
This prosodic pattern is characterized with two (typically elongated) regions of flat pitch with a short-intensity dip. According to Ward \cite{ward2019prosodic}, this construct generally functions as a cue for the listener to take action under specific conditions: (i) when there is a single clearly appropriate action; (ii) when this action is required (e.g., by a social norm); (iii) when the action is simple to execute; and (iv) when it should be carried out immediately. In our qualitative analysis, we noted evidence for three distinct variations of this prosodic pattern, each serving a different pragmatic function.

\subsubsection{Desist}
This variation of Minor Third construction often features a downstepped `stop', effectively cueing the cessation of an action \cite{ward2019prosodic}. This pattern can be compelling even without words, as demonstrated when discouraging a toddler from reaching for cookies before snack time with a simple `\textit{\underline{uh oh}}'. An example of this construct in our study is a participant commanding \textit{`keep going...'} as the robot was moving towards the ball, followed by a harsh down-stepped \textit{\underline{`stop!}'} as it critically approaches the wall.

\subsubsection{Calling}
A variation of minor third construct is frequently used for calling someone, to the extent that prior work refers to it as as the ``Calling Contour'' \cite{ladd1978stylized}. An example of this from our study is an interaction wherein a participant used this construct to call the robot using a two-syllable word in place of a name, `\textit{\underline{dog-go}}' (elongated, with the first syllable at a higher pitch), to get the robot to turn and direct its attention to the speaker, from across the room.

\subsubsection{Reprimand}
Another variant of Minor Third construct uses superimposed clipped ends to cue a strong reprimand, with glottal stops after each syllable turning the generic action-cueing effect into a more controlling one \cite{ward2019prosodic}, such as in saying \textit{`\underline{bad dog}'}!

\subsection{Backchannelling Construction}
This construct is characterized by its lengthened, quiet utterances with a typically flat pitch and slightly creaky voice. Its primary function is to `encourage the continuation of an action'. It is often used in communication to establish shared knowledge or to serve as a subtle cue for continuation \cite{ward2019prosodic}. An interaction instance in our study that exemplifies this construct is a participant saying \textit{\underline{'nice... nice...'}}, rhythmically and calmly, to encourage the robot to keep walking in the same manner, speed and direction. 

\subsection{Positive Assessment Construction}
This construct is characterized by a region of a relatively high pitch, then a region of increased loudness with clear voicing, and finally a clipped ending marked by a sharp drop in intensity. Its main function is to `express positive assessment' \cite{ward2019prosodic}. An example of this in our study was an interaction instance in which after the robot successfully completed the task, the participant patted its back and used this construct to utter \textit{`\underline{good \# boy'}}!

\section{Discussion}
\label{discussion}
Our study focused on exploring intuitive robotic interfaces through a Research through Design approach. Participants instructed a quadruped robot in an obstacle course, with their interactions recorded. Our observations revealed that participants intuitively relied on various prosodic constructs to convey key information. These included (a) the interpolation construct to signal urgency, (b) the minor third construct for enforcing immediate cessation, calling, and indicating reprimand, (c) the backchannelling construct to encourage the continuation of an action, and (d) the positive assessment construct for providing positive reinforcement.

Prosody not only facilitated the communication of time-sensitive commands in our study but also played a crucial role in disambiguating communicative signals. A notable example, discussed earlier in the paper, is the varied use of the word \textit{`nice'}-- This word was sometimes used to indicate \textit{`keep going'}, and at other times, \textit{`stop'}. Our analysis of prosodic cues revealed that in instances where \textit{'nice'} meant \textit{`keep going'}, the backchannelling construct was employed; conversely, when it was used to signal \textit{`stop'}, it was in using the minor third (desist) construct. This highlights how prosody can provide vital context to seemingly straightforward lexical terms.

While in this study, interpretation of prosodic cues was done via a human wizard, features such as fundamental frequency (F0), loudness, sound and silence duration and other spectral information can be \textit{objectively} extracted using frame-by-frame acoustic analysis of real-time spoken interaction, captured on a live mic. Given an annotated dataset, such prosodic features can then be fed into models powerful enough to detect temporal dependencies between them. As discussed earlier, computational systems using basic sets of prosodic features paired with classfiers may outperform humans in detection for many tasks. For future work exploring computational prosody, we recommend considering openSMILE~\cite{eyben2010opensmile} (an open-source toolkit) and this starter bibliography by Ward \cite{ward2021computational}.

Furthermore, gaining a deeper understanding of prosody is pragmatic uses not just for robots to make sense of human communication, but also for them to generate communicative signals that people can intuitively understand. As discussed earlier, specific prosodic constructs can indicate either positive assessment, or reprimand. A robot detecting a prosodic `reprimand' cue, can interpret it as in-context punitive feedback and use it for adjustment or cessation of an action, effectively leveraging prosody for reinforcement learning. In addition, the robot can proceed to generate prosodic cues to convey a ``sincere" apology, for example by employing a non-verbal tone that matches the \textit{`giving in'} construct (a prosodic configuration that conveys \textit{sincerity in an apology}, which children learn to use early in life. For details, see \cite{ward2019prosodic}). That is, communicative nuances embedded in prosody can be used to enhance a robot's ability to learn and respond appropriately and empathetically.

A key advantage of prosody lies in its gradational nature, which allows us to not only communicate cues such as urgency, but also discern the relative intensity expressed in those cues. That is, we can listen to two lexically-identical utterances and understand which one \textit{sounds more urgent}. In other words, prosody is `a matter of degree', and humans have fine control over its subtle variations in speech \cite{ward19_interspeech}, which allows for nuanced control. It is worth noting that by applying a low-pass filter to speech, it is possible to only preserve prosodic elements, discarding verbal content while still enabling meaningful communication.

Computational prosody for HRI presents significant challenges. Disentangling prosody from lexical content, while extracting cues relevant to robot actions necessitates sophisticated approaches. Furthermore, integrating these prosodic cues with lexical content in a way that maintains contextual understanding introduces further complexities.
Adding proficiency in non-verbal modalities can improve the robustness of future intuitive robotic interfaces. To achieve this, a deeper scientific understanding of prosody with an emphasis on `cross-modality integration' is needed.

Finally, we argue that prosody lends itself well to designing control interfaces for for mobile agents, because it mimics how we communicate with animals-- not using our words, but using our prosody \cite{gergely2023dog, korcsok2020artificial}. In fact, work on evolutionary biology suggests that ``affective prosody in human acoustic communication has deep-reaching phylogenetic roots, deriving from precursors already present and relevant in the vocal communication systems of nonhuman mammals'' \cite{zimmermann2013toward}. Throughout history, humans have partnered with animals (non-human intelligent mobile agents) to extend their own embodiment and do things they could not do on their own \cite{darling2021new}. We have trained animals to be receptive to our communication and learn from us, by devising basic sets of `controller commands' (e.g., American Kennel Club (AKC) recommends 5 basic commands for training puppies \cite{Lunchick_2023}). Looking ahead, future research could take inspiration and investigate how prosodic cues might be leveraged for designing basic set of robot control commands.

\vspace{-2mm}
\section{Conclusion}
The study explores intuitive robotic interfaces using a Research through Design approach. Participants interacted with a quadruped robot on an obstacle course, using natural interaction to convey instructions. A key finding includes evidence for intuitive use of prosody for signaling urgency, stopping, encouragement, punitive and positive feedback. We highlight the role of prosody in communicating time-sensitive commands and disambiguating communicative signals, and emphasize its importance in adaptive robotic learning and communication. This work underscores the potential of prosody for designing future intuitive robotic control interfaces and enabling continual learning and personalization in HRI. 

\bibliographystyle{ACM-Reference-Format}
\bibliography{main}

\end{document}